\documentclass[conference,letterpaper]{IEEEtran}

\usepackage[utf8]{inputenc} 
\usepackage[T1]{fontenc}
\usepackage{url}
\usepackage{ifthen}
\usepackage{cite}
\usepackage[cmex10]{amsmath} 
\usepackage{enumitem}

\usepackage{amsmath,amssymb,amsfonts}
\usepackage{algorithmic}
\usepackage{graphicx}
\usepackage{textcomp}
\usepackage{xcolor}
\usepackage{bbm}
\usepackage[ruled,vlined]{algorithm2e}

\DeclareMathOperator{\Ex}{\mathbb{E}}
\DeclareMathOperator{\Prob}{\mathbb{P}}
\DeclareMathOperator{\F}{\mathcal{F}}
\DeclareMathOperator{\Hp}{\mathcal{H}}

\DeclareMathOperator*{\argmax}{arg\,max}

\newtheorem{theorem}{Theorem}
\newtheorem{lemma}[theorem]{Lemma}
\newtheorem{proposition}[theorem]{Proposition}

\def\BibTeX{{\rm B\kern-.05em{\sc i\kern-.025em b}\kern-.08em
		T\kern-.1667em\lower.7ex\hbox{E}\kern-.125emX}}

\begin{document}
	\title{Asymptotic Performance of Thompson Sampling in the Batched Multi-Armed Bandits\\
		\thanks{
			This work was partly supported by NSF award NeTS 1817205 and the Center for Science of Information (CSoI), an NSF Science and Technology Center, under grant agreement CCF-0939370. 
	}} 
	

	\author{\IEEEauthorblockN{Cem Kalkanl\text{\i}  and Ayfer \"{O}zg\"{u}r}
		\IEEEauthorblockA{Department of Electrical Engineering \\
			Stanford University\\
			Email: \{cemk, aozgur\}@stanford.edu}
	}

	
	\IEEEoverridecommandlockouts
	\maketitle
	
\begin{abstract}
	We study the asymptotic performance of the Thompson sampling algorithm 
	in the batched multi-armed bandit setting where the time horizon $T$ is divided into batches, and the agent is not able to observe the rewards of her actions until the end of each batch. We show that in this batched setting, Thompson sampling achieves the same asymptotic performance as in the case where instantaneous feedback is available after each action, provided that the batch sizes increase subexponentially. This result implies that Thompson sampling can maintain its performance even if it receives delayed feedback in $\omega(\log T)$ batches. We further propose an adaptive batching scheme that reduces the number of batches to $\Theta(\log T)$ while maintaining the same performance.  Although  the batched multi-armed bandit setting  has been considered in several recent works, previous results  rely on tailored algorithms for the batched setting, which optimize the batch structure and prioritize exploration in the beginning of the experiment to eliminate suboptimal actions. We show that Thompson sampling, on the other hand, is able to achieve a similar asymptotic performance in the batched setting without any modifications. 
\end{abstract}

	
	\section{INTRODUCTION}

	The multi-armed bandit problem models the relationship between learning new information, i.e. exploration, and using the current knowledge to maximize rewards, i.e. exploitation, in sequential decision problems \cite{robbins1952some}. In this setting, the agent, whose goal is to accumulate as much reward as possible, repeatedly selects actions using her estimates about the system, and updates these estimates once she receives feedback, e.g a reward. Although it is ideal for the agent to receive instantaneous feedback so that she can adjust her algorithm before the next action instance, many real-world settings limit the number of interactions the agent can have with the system. For example, in medical applications \cite{thompson1933likelihood}, treatments are run in parallel for groups of patients, and the experimenter has to wait for the outcome of one trial before designing the next set of clinical trials. In online advertising \cite{schwartz2017customer}, it is not practical to update the algorithm every time a user generates feedback since there may be millions of responses per second. 
	
	Perchet et al. \cite{perchet2016batched} modeled this problem as the batched multi-armed bandit. Here, the duration of the experiment $T$ is divided into $M$ batches, and the agent is unable to adjust her algorithm or receive any feedback until the end of each batch. For the batched two-armed bandit in which each pull of an arm produces an unknown deterministic reward corrupted by a sub-Gaussian noise, i.e. the frequentist setting, they proposed an explore-then-commit (ETC) policy, where the agent plays both arms the same number of times until the terminal batch and commits to the better performing arm in the last round unless the sample mean of one arm sufficiently dominates the other in earlier batches. They show that this algorithm achieves the optimal problem-dependent regret $O(\log(T))$ matching the asymptotic lower bound $\Omega(\log(T))$  in \cite{lai1985asymptotically} for the non-batched case (i.e. $M=T$) with only $O(\log(T/\log(T)))$batches. Gao et al. \cite{gao2019batched} later proposed a batched successive elimination (BaSE) policy for batched multi-armed bandits with arbitrary (finite) number of arms. The BaSE algorithm is similar to the ETC algorithm in that in each batch the agent plays each of the actions in a set of remaining actions in a round robin fashion, and eliminates the underperforming arms at the end of each batch. It is shown that this policy achieves a problem-dependent regret $O(\log(T))$ with $O(\log(T))$ batches when the batch sizes increase exponentially. More recently, several other optimal batched algorithms have appeared in \cite{esfandiari2019batched,jin2020double}. These algorithms share the same explore first-exploit later principle as ETC and BaSE with slight modifications, such as allowing the agent to commit to the best performing arm earlier or having a small exploit stage in the middle of the exploration phase.  
	
	In this paper, we study the cumulative random regret of the Thompson sampling algorithm \cite{thompson1933likelihood} in the batched multi-armed bandits problem. In the Thompson sampling algorithm, the agent chooses an action randomly according to its likelihood of being optimal,
	and after receiving feedback, i.e. observing rewards, updates its beliefs about the optimal action. Thompson sampling has been successfully applied to a broad range of online optimization problems \cite{chapelle2011empirical,schwartz2017customer} and has been thoroughly analyzed in both the frequentist \cite{kaufmann2012thompson,korda2013thompson,abeille2017linear,agrawal2017near} and the Bayesian settings \cite{russo2014learning,russo2016information}.

	In this paper, we ask whether Thompson sampling, a strategy that naturally balances between exploration and exploitation at each step, can perform well in the batched setting. Equivalently, can Thompson sampling maintain its performance when feedback is delayed and the agent is allowed to update its beliefs only at the end of several batches? The earlier algorithms developed for the batched setting \cite{perchet2016batched,gao2019batched} suggest that batched algorithms need to heavily prioritize exploration in the initial batches. This is because when the batch sizes increase exponentially, the cumulative regret is dominated by the contribution from the final batches, and more aggressive exploration is needed to eliminate the possibility that a suboptimal arm is played in the final batch. It is unclear whether a strategy like Thompson sampling can achieve a good performance in this setting, without explicitly prioritizing exploration.

	In this paper, we answer this question in the affirmative. We consider the frequentist convention of \cite{agrawal2017near}, where Thompson sampling is operated with Gaussian priors, and  show that Thompson sampling retains the same asymptotic regret of order $O(\log(T))$ as in the case where the agent receives instantaneous feedback after each action, for any batch structure as long as there are $\omega(\log(T))$ batches. Notably, this excludes the case where the batch sizes increase exponentially and the number of batches is $\Theta(\log T)$. We show that the same performance can be achieved with $O(\log T)$ batches if the Thompson sampling agent is allowed to adaptively choose its batch sizes. We propose a simple adaptive batching scheme called iPASE (inverse ProbAbility batch SizE), where at the end of each batch the Thompson sampling agent looks at the arm selection probabilities (recall that with Thompson sampling the selection probability of an arm corresponds to its likelihood of being optimal) and chooses the size of the next batch as the multiplicative inverse of the second largest of these probabilities. This allows the algorithm to naturally pace itself; as the selection probability of one arm dominates the others, the batches become larger and larger. We show that with this adaptive batching scheme Thompson sampling achieves $O(\log(T))$ regret, while total number of batches can be bounded by $O(\log(T))$. 

	\section{PRELIMINARIES}\label{setting}
	
	\subsection{Batched Multi-Armed Bandit}\label{bab}
	We consider the batched multi-armed bandit setting. Here we have $I\in\mathbb{Z}^+$ many arms, where each consecutive pull of the $i^{th}$ arm produces i.i.d. rewards $\{Y_{i,t}\}_{t=1}^\infty$ such that
	\begin{align*}
		\Ex[Y_{i,1}]&=\mu_i\in\mathbb{R},\\
		\Ex[(Y_{i,1}-\mu_i)^2]&=\sigma_i^2<\infty.
	\end{align*}
	These $\{\mu_i\}$ are assumed to be unknown deterministic parameters. In this model, there is an agent, whose goal is to accumulate as much reward as possible by repeatedly pulling these arms. Therefore, at each time instance $t$, the agent plays an arm $A_t\in\{1,2,...,I\}$ and receives the reward $Y_{A_t,t}$. Since she can only act causally and does not know $\{\mu_i\}$, she can only use the past observations, $\Hp_t=\{A_1,Y_{A_t,1},...,A_t,Y_{A_t,t}\}$ where $\Hp_{0}=\emptyset$, to select the next action $A_{t+1}$.
	
	We consider the batched version of this problem, where the feedback the agent receives in the form of rewards may not be synchronized with the action instances $t$. In other words, there are batch end points $0=T_0<T_1<...$, and the actions the agent plays in the batch cycle of $[T_{j-1}+1,T_j]$ can depend only on the information presented in $\Hp_{T_{j-1}}$ for any $j\in\mathbb{Z}^+$. This condition causes the probability of the agent selecting the $i^{th}$ arm to obey the following rule for any $t$
	\begin{equation*}
		\Prob(A_{t}=i|\Hp_{t-1})=\Prob(A_t=i|\Hp_{T_{b(t)}}),
	\end{equation*} 
	where $b(t)=\max\{j\in\mathbb{Z}_{\geq 0}|t-1\geq T_j\}$. Similarly, we define $B(t)=\max\{j\in\mathbb{Z}_{\geq 0}|t\geq T_j\}$, which denotes the number of times the agent has received batched feedback from the system or the number of the completed batch cycles.  
	
	In the sequel, we will consider the above problem under two different assumptions on how the batch structure is chosen:
	\begin{itemize}
		\item Adversarial batching: $T_j$ is chosen by an adversary at time $T_{j-1}$ upon observing $\Hp_{T_{j-1}}$, and revealed to the agent; 
		\item Adaptive batching: $T_j$ is chosen by the agent at time $T_{j-1}$ upon observing $\Hp_{T_{j-1}}$.
	\end{itemize}
	Note that in both cases $T_j$ can be chosen as a function of $\Hp_{T_{j-1}}$. However, adversarial batching accounts for the worst case choice of the batching structure, while adaptive batching corresponds to the optimistic case where the batch sizes can be adopted to the history of the process. In addition, we assume that the agent receives feedback from the system infinitely often, i.e $b(t)\rightarrow\infty$, similarly $B(t)\rightarrow\infty$, almost surely as $t\rightarrow\infty$.
	
	In this setting, we let $\mu_1>\mu_i$ for $i\geq 2$. Given that the agent aims to maximize her cumulative reward, she would only play the $1^{st}$ arm if she knew the hidden system parameters $\{\mu_i\}$. This observation naturally leads to the cumulative regret term, $R(T)$:
	\begin{align}
		&R(T)=\sum_{t=1}^T Y_{1,t}-Y_{A_t,t}=\sum_{i=2}^{I}\sum_{t=1}^{T}\mathbbm{1}_{\{A_t=i\}}(Y_{1,t}-Y_{i,t})\label{deee1}\\
		&=\Big(\sum_{i=2}^{I}\sum_{t=1}^{T}\mathbbm{1}_{\{A_t=i\}}(Y_{1,t}-\mu_1+\mu_i-Y_{i,t})\Big)+\Big(\sum_{i=2}^{I}\Delta_iN_i(T)\Big),\label{f1}
	\end{align}
	where $\mu_1-\mu_i=\Delta_i$ and $N_{i}(T)=\sum_{t=1}^{T}\mathbbm{1}_{\{A_t=i\}}$ for any $i$. In view of \eqref{deee1}, we also define the total regret accumulated by the pulls of the arm $i$, $R_i(T)$:
	\begin{equation*}
		R_i(T)=\sum_{t=1}^{T}\mathbbm{1}_{\{A_t=i\}}(Y_{1,t}-Y_{i,t}).
	\end{equation*}
	As a result, \eqref{deee1} can also be expressed as $R(T)=\sum_{i=2}^I R_i(T)$.
	
	Finally, we let $S_i(T)$ be the total measurement effort allocated to the $i^{th}$ arm such that
	\begin{equation*}
		S_i(T)=\sum_{t=1}^{T}\Prob(A_t=i|\Hp_{t-1}).
	\end{equation*}
	
	\subsection{Thompson Sampling}
	
	In this paper, the agent employs the Thompson sampling algorithm, and to operate it, she presumes that the system parameters, $\{\mu_i\}$, and the noise terms, $\{Y_{i,t}-\mu_i\}$, are mutually independent standard normal random variables. Thus, the agent plays an arm according to its likelihood of being optimal under this assumption. In accordance with Algorithm \ref{algo}, we let $\hat{\mu}_i(T)$ and $\hat{\sigma}^2_i(T)$ for any $T\in\mathbb{Z}^+$ be
	\begin{equation*}
		\hat{\mu}_i(T)=\frac{\sum_{t=1}^T \mathbbm{1}_{\{A_t=i\}}Y_{i,t}}{1+N_i(T)}\text{ and }\hat{\sigma}^2_i(T)=\frac{1}{1+N_i(T)},
	\end{equation*} 
	where $\hat{\mu}_i(T)=0$ and $\hat{\sigma}^2_i(T)=1$ if $N_i(T)=0$ or $T=0$. 
	
	The implementation of Thompson sampling for the batched multi-armed bandits is shown in Algorithm \ref{algo}. It is easy to check that the pseudo posterior distributions used in the algorithm correspond to the standard normal distribution priors and reward distributions. Note that Algorithm \ref{algo} also accommodates the case where the agent adaptively selects the batch sizes. This algorithm simplifies to the Algorithm 2 of \cite{agrawal2017near}, called ``Thompson Sampling Using Gaussian Priors", when $T_j=j$ for any $j\in\mathbb{Z}^+$.
	
	\begin{algorithm}[h]\label{algo}
		\SetAlgoLined
		\textbf{Input:} The first batch size $T_1$\\
		\textbf{Initialization:} $j=1$, $\hat{\mu}_i(0)=0$, $\hat{\sigma}^2_i(0)=1$, $N_i(0)=0$.\\
		\While{Experiment Run}{
			\textbf{Perform the $j^{th}$ Batch Cycle Operations:}
			\For{$t\leftarrow T_{j-1}+1$ \textbf{to} $T_j$}{
				\textbf{Sample for Each Arm:} $\theta_i(t)\sim\mathcal{N}(\hat{\mu}_i(T_{j-1}),\hat{\sigma}^2_i(T_{j-1}))$\\
				\textbf{Play an Arm:} $A_t=\argmax_{i} \theta_i(t)$\\
				\textbf{Update the Pull Count:} $N_{A_t}(t)\gets N_{A_t}(t-1)+1$
			}		
			\textbf{Receive the Rewards:} $\{Y_{A_t,t}\}_{t=T_{j-1}+1}^{T_j}$\\
			\textbf{Update the Statistics for Each Arm:}\\
			\eIf{$N_i(T_{j-1})=N_i(T_j)$}{
				$\hat{\mu}_i(T_j)\gets\hat{\mu}_i(T_{j-1})$\\
				$\hat{\sigma}^2_i(T_j)\gets\hat{\sigma}^2_i(T_{j-1})$	
			}{
				$\hat{\mu}_i(T_j)\gets\frac{(1+N_i(T_{j-1}))\hat{\mu}_i(T_{j-1})+\sum_{t=T_{j-1}+1}^{T_j}\mathbbm{1}_{\{A_t=i\}}Y_{i,t}}{1+N_i(T_j)}$\\
				$\hat{\sigma}^2(T_j)\gets\frac{1}{1+N_i(T_j)}$
			}
			\textbf{Select the Size of the Next Batch:} $T_{j+1}-T_{j}$ (can be a predetermined constant)\\
			\textbf{Update the Batch Index:} $j\gets j+1$\\
		}
		\caption{Thompson Sampling}
	\end{algorithm}

	Notice that for any $t_1$ and $t_2$ such that $T_{j-1}<t_1,t_2\leq T_{j}$ for some $j\in\mathbb{Z}^+$, we have
	\begin{equation}
		\Prob(A_{t_1}=i|\Hp_{T_{j-1}})=\Prob(A_{t_2}=i|\Hp_{T_{j-1}})\label{cond1}
	\end{equation}
	with the Thompson sampling algorithm. This equality follows from the fact that the sampling process is fixed during the same batch cycle even if the batch sizes are chosen adaptively given the history. Note that since $T_j$ is adapted to $\Hp_{T_{j-1}}$, conditioned on $\Hp_{T_{j-1}}$, $T_j$ and the sampling process of the $j^{th}$ batch, i.e. the random variables associated with the sampling process, are independent. As a result of \eqref{cond1}, the total measurement effort assigned to the $i^{th}$ arm in the first $j$ batch cycles, namely $S_i(T_j)$, can be rewritten as
	\begin{equation}
		S_i(T_j)=\sum_{m=1}^j (T_{m}-T_{m-1})\Prob(A_{T_{m-1}+1}=i|\Hp_{T_{m-1}}),\label{maineq}
	\end{equation} 
	or as
	\begin{equation}
		S_i(T_j)=(T_{j}-T_{j-1})\Prob(A_{T_{j-1}+1}=i|\Hp_{T_{j-1}})+S_i(T_{j-1}).\label{sit1}
	\end{equation}

	\section{MAIN RESULTS}
	In this section, we state the main results of our paper.
	
	\subsection{ A Regret Bound For Adversarial Batch Structures}

	We have the following result for adversarial batching.
	
	\begin{theorem}\label{thm5}
		Given that the adversarially chosen batching scheme satisfy 
		\begin{equation}
			\Prob\Bigg(\limsup_{j\rightarrow\infty}\log_{T_j}(T_{j+1}-T_j)<1\Bigg)
			=1,\label{ggg3}
		\end{equation}
		the cumulative regret, $R(T)$, of the Thompson sampling agent satisfies the following limit almost surely:
		\begin{equation*}
			\lim_{T\rightarrow\infty}\frac{R(T)}{\log(T)}=\sum_{i=2 }^I\frac{2}{\Delta_i}.
		\end{equation*}
	\end{theorem}
	Theorem~\ref{thm5} states that if the batching structure $\{T_1, T_2, \dots\}$ generated by the adversary satisfies \eqref{ggg3}, then asymptotic regret of Thompson sampling is bounded by $O(\log T)$. In particular, the theorem holds if the batching structure is fixed and given ahead of time, in which case the condition  \eqref{ggg3} reduces to $\limsup_{j\rightarrow\infty}\log_{T_j}(T_{j+1}-T_j)<1$. This condition requires the batch sizes to have a subexponential growth rate. Note, for example, that the condition will be satisfied if $T_j$'s increase polynomially, i.e. $T_j=j^p$ for any fixed $p>0$, while an exponential growth rate would violate the condition.


	Theorem \ref{thm5} is the first result in the literature that provides a theoretical guarantee for the performance of Thompson sampling in the batched multi-armed bandit setting. Note that Theorem \ref{thm5} also applies in the case when instantaneous feedback is available, i.e. $T_j=j$, hence it implies that the performance of Thompson sampling remains the same as long the batch size has a subexponential growth rate. 
	
	
	
	We contrast our result with Korda et al. \cite{korda2013thompson}, which provides an asymptotic problem-dependent regret bound for Thompson sampling in the classical bandit setup. They show  that in this case, the expected regret of Thompson sampling, $\Ex[R(T)]$, satisfies
	\begin{equation}
		\lim_{T\rightarrow\infty}\frac{\Ex[R(T)]}{\log(T)}\leq\sum_{i=2}^I\frac{\Delta_i}{KL(i,1)},\label{ggg4}
	\end{equation}
	if Thompson sampling is operated with Jeffrey's prior and the reward distributions are from 1-dimensional exponential family. Here $KL(i,1)$ is the Kullback-Leibler divergence between the distributions of $Y_{i,t}$ and $Y_{1,t}$. We first note that both Theorem~\ref{thm5} and \eqref{ggg4} provide the same convergence rate when the rewards are corrupted by a standard normal noise, i.e. $Y_{i,t}\sim\mathcal{N}(\mu_i,1)$. Note that a simple computation shows $KL(i,1)=\Delta_i^2/2$. However, notice the scope of the two results are different in the sense that \eqref{ggg4} bounds the regret of Thompson sampling when the reward distributions are known and the Thompson sampling algorithm samples its actions from the corresponding posterior distribution. As a result, their bound has an explicit dependence on the reward distributions through the  Kullback-Leibler divergence. In contrast, we provide an upper bound on the asymptotic regret of Thompson sampling when  the reward distributions are not known and the posterior distribution is computed by assuming Gaussian reward distributions, a technique first analyzed in \cite{agrawal2017near}.

	We next introduce an adaptive batching scheme that achieves an asymptotic scaling of order $\log(T)$ in both the cumulative random regret $R(T)$ and the number of completed batch cycles $B(T)$. 
	

	\subsection{Inverse Probability Batch Size (iPASE)}
	
	We next propose a simple strategy, which we call iPASE, for the agent to adaptively choose the batch sizes. In this scheme, the agent uses the following rule to decide on the next batch size:
	\begin{equation}
		T_{j+1}-T_j=\Big\lfloor\frac{1}{\Prob_{2}(T_j)}\Big\rfloor.\label{inv}
	\end{equation}  
	Here $\Prob_{2}(T_j)$ is the second biggest element of the set $\{\Prob(A_{T_j+1}=i|\Hp_{T_{j}})\}_{i=1}^I$, and $\lfloor\cdot\rfloor$ is the floor function. As can be seen from \eqref{inv}, iPASE follows a simple evolution rule, and since $\theta_i$s are Gaussian random variables conditioned on the past observations, \eqref{inv} can be calculated with numerical integration. Note that it can also be approximated with the Monte Carlo method.
	
	iPASE is motivated by the following simple idea. Let's assume that the agent is allowed to choose the batch sizes adaptively and the probability of the agent choosing any arm before deciding on the batch size is around $1/I$. That means the agent is not confident in her estimates and naturally should keep the size of the next batch small. On the contrary, if the probability of selecting one arm dominates the others', then the size of the next batch should be comparatively big. Although the agent's estimates may be wrong at the beginning of the experiment, the probability of selecting the optimal arm should dominate the other probabilities as the algorithm evolves. We next show that iPASE achieve the asymptotic performance in Theorem 1 with only $O(\log(T))$ batches

	\begin{theorem}\label{thm6}
		The cumulative regret, $R(T)$, and the number of completed batch cycles, $B(T)$, of Thompson sampling with iPASE satisfy the following  inequalities almost surely:
		\begin{equation*}
			\limsup_{T\rightarrow\infty}\frac{R(T)}{\log(T)}\leq\sum_{i=2}^I\frac{2}{\Delta_i},
		\end{equation*}
		and
		\begin{equation*}
			\limsup_{T\rightarrow\infty}\frac{B(T)}{\log(T)}\leq\sum_{i=2}^I\frac{2}{\Delta_i^2}.
		\end{equation*} 
	\end{theorem}
	We provide the proof of this theorem in the case of two-armed bandits at the end of this paper.
	
	The combination of Theorems \ref{thm5} and \ref{thm6} shows that Thompson sampling with iPASE achieves the same asymptotic performance as Thompson sampling in the classical bandit setup with only $O(\log(T))$ number of batches; consequently, batch complexity of iPASE asymptotically matches the batch complexity of the algorithms proposed by Gao et al. \cite{gao2019batched} and Esfandiari et al. \cite{esfandiari2019batched} in the case where the expected rewards of each arm, $\{\mu_i\}$, are fixed unknown constants, i.e. not functions of $T$. On the other hand, the asymptotically optimal DETC algorithm of \cite{jin2020double} requires only $O(1)$ expected number of batches. However, unlike the algorithms of \cite{gao2019batched,esfandiari2019batched} and iPASE, DETC does not guarantee that the number of batches won't exceed the order of $\log(T)$ and it is limited to two-armed bandits.  
	
	\section{EXPERIMENTS}\label{EXP}
\begin{figure}[h]
		\centering
		
		\includegraphics[scale=0.45]{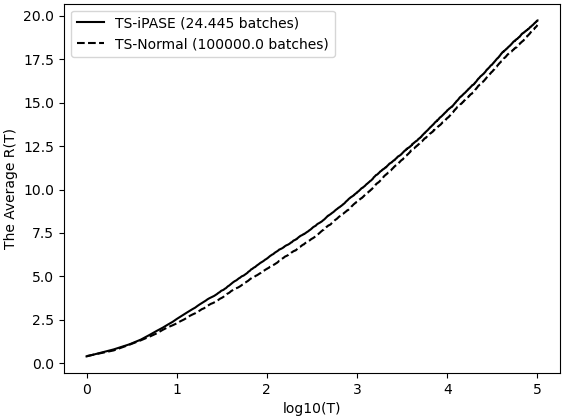}

		\caption{Average regret vs. the time horizon T for $Y_{1,t} \sim \text{Bern} (0.9)$ and $Y_{2,t} \sim \text{Bern} (0.1)$}
		\label{fig1}
	\end{figure}

	We next compare the numerical performance of Thompson sampling without any batching (TS-Normal) and Thompson samling with iPASE (TS-iPASE). In Figure \ref{fig1}, we consider $T=10^5$ and $I=2$ with Bernoulli reward distributions. Here TS-iPASE is implemented with the Monte Carlo method and the following figure is the result of an experiment averaged over 400 repeats. Finally, we record the average number of batches each algorithm used throughout the experiment in the parenthesis to the right of the algorithm names on the figure. Our results indicate that iPASE can dramatically decrease the feedback required by Thompson sampling without impacting its performance.

	\section{Proof of Theorem \ref{thm6} for Two-Armed Bandits}\label{2armed}
	
	Here, we prove Theorem \ref{thm6} in the case of two-armed bandits. Before we start, we first state three essential results that apply to Thompson sampling with any batching scheme as long as the conditions in Section \ref{setting} are satisfied. In particular, these results are true in the case of iPASE (see Appendix Section E for the proof of this statement).  We now present the three important results. The proofs of these are mostly technical and given in Appendix Section C. 
	
	\begin{enumerate}[label=(\roman*)]
		\item Conditional probability of selecting the $i^{th}$ arm in the $j+1^{th}$ batch, i.e. $\Prob(A_{T_j+1}=i|\Hp_{T_{j}})$, almost surely converges to 1 or 0 depending on $i$ being 1 or not respectively.
	\end{enumerate}
	This result shows that Thompson sampling accurately detects whether an action is optimal or suboptimal as the batch count $j$ increases.
	\begin{enumerate}[label=(\roman*)]
		\setcounter{enumi}{1}
		\item $S_i(T_j)$ almost surely diverges for all arms and 
		\begin{equation*}
			\lim_{T\rightarrow\infty}\frac{N_i(T)}{S_i(T)}=1,\quad\lim_{T\rightarrow\infty}\frac{R_i(T)}{S_i(T)}=\Delta_i
		\end{equation*}
		with probability 1 for $i\geq 2$.
	\end{enumerate}
	(ii) indicates that Thompson sampling allocates infinite amount of measurement effort, $S_i(T)$, to all actions throughout the experiment. In addition, the relationship between $S_i(T)$ and $R_i(T)$ suggests that analyzing $S_i(T)$ is enough to characterize the asymptotic behavior of $R(T)$.
	\begin{enumerate}[label=(\roman*)]
		\setcounter{enumi}{2}
		
		\item \begin{proposition}\label{cor2}
			Suppose the setting is as described in Section \ref{setting}, then for any $i\geq 2$
			\begin{equation*}
				\lim_{j\rightarrow\infty}\frac{-\log(\Prob(A_{T_j+1}=i|\Hp_{T_{j}}))}{S_i(T_{j})}=\frac{\Delta_i^2}{2}
			\end{equation*}
			almost surely.
		\end{proposition} 
	\end{enumerate}
	In view of \eqref{sit1}, Proposition \ref{cor2} describes how $S_i(T_j)$ evolves as the batch count $j$ increases, and it is the main ingredient in our proof. As such, we first provide the proof idea of this proposition in the two-armed bandits setup. In this setting, the form of $\Prob(A_{T_j+1}=2|\Hp_{T_{j}})$ simplifies to the following:
	\begin{align}
		\Prob(A_{T_j+1}=2|\Hp_{T_{j}})&=\Prob(\theta_2(T_j+1)\geq\theta_1(T_j+1)|\Hp_{T_{j}})\nonumber\\
		&=Q\Big(\frac{\hat{\mu}_1(T_{j})-\hat{\mu}_2(T_{j})}{\sqrt{\hat{\sigma}_1^2(T_{j})+\hat{\sigma}_2^2(T_{j})}}\Big),\label{kkkkkkk1}
	\end{align}
	where $Q(\delta)=\Prob(X\geq\delta)$ for any standard normal random variable $X$. The last equality follows from the fact that conditioned on the past $\Hp_{T_j}$, $\theta_i$s are Gaussian random variables. Here, considering that Thompson sampling accurately detects the optimal action by (i), we can also expect it to predict the expected reward of each action asymptotically. This observation means that $\hat{\mu}_1(T_{j})-\hat{\mu}_2(T_{j})$ should almost surely converge to $\Delta_2$. Similarly, the combination of (i) and (ii) suggests that the number of times the Thompson sampling agent picks the first arm, $N_1(T)$, should dominate the second arm's, $N_2(T)$. This conclusion results in $\hat{\sigma}_1^2(T_{j})$ being significantly smaller than $\hat{\sigma}_2^2(T_{j})$ for large $j$ by the definition of $\hat{\sigma}_i^2(T_{j})$s. As a result, the preceding analysis shows that the term inside of $Q(\cdot)$ in \eqref{kkkkkkk1} diverges at the same rate as $\Delta_2\sqrt{N_2(T_j)}$ as $j\rightarrow\infty$. However, by the definition of the function $Q(\cdot)$, we know that $Q(\delta)$ can be approximated by $\text{poly}(\delta)\exp(-\delta^2/2)$ as $\delta$ diverges to infinity. This fact leads to $-\log(Q(\delta))$ having the same rate as $\delta^2/2$ when $\delta\rightarrow\infty$. In view of \eqref{kkkkkkk1} and (ii), replacing $\delta$ with $\Delta_2\sqrt{N_2(T_j)}$ finishes the proof overview of Proposition \ref{cor2}. The technical proof of this proposition formalizes the preceding set of arguments. 
	
	
	We are now ready to prove Theorem \ref{thm6}. As can be seen from (i), there exists an almost surely finite random $K$ such that if $k\geq K$, then $\Prob_{2}(T_k)=\Prob(A_{T_k+1}=2|\Hp_{T_{k}})$ by the definition of $\Prob_{2}(T_k)$. This observation leads to the following set of inequalities:
	\begin{align}
		&\frac{S_2(T_{k+1})}{\log(T_{k+1})}\leq\frac{S_2(T_k)+\Prob(A_{T_k+1}=2|\Hp_{T_{k}})\Big\lfloor\frac{1}{\Prob(A_{T_k+1}=2|\Hp_{T_{k}})}\Big\rfloor}{\log\Big(\frac{1}{\Prob(A_{T_k+1}=2|\Hp_{T_{k}})}-1\Big)}\label{mm1}\\
		&\leq\frac{S_2(T_k)+1}{\log\Big(\frac{1}{\Prob(A_{T_k+1}=2|\Hp_{T_{k}})}-1\Big)}=\frac{S_2(T_k)+1}{S_2(T_k)}\nonumber\\
		&\times\frac{S_2(T_k)}{\log\Big(\frac{1}{\Prob(A_{T_k+1}=2|\Hp_{T_{k}})}\Big)}\times\frac{\log\Big(\frac{1}{\Prob(A_{T_k+1}=2|\Hp_{T_{k}})}\Big)}{\log\Big(\frac{1}{\Prob(A_{T_k+1}=2|\Hp_{T_{k}})}-1\Big)}\label{mm2}
	\end{align}
	if $k\geq K$. Here, \eqref{mm1} follows from \eqref{sit1} and the batch size rule \eqref{inv} of iPASE. Now, we only need to analyze the three multiplicand fractions in \eqref{mm2}. The first fraction almost surely converges to 1 by (ii), the second fraction almost surely converges to $2/\Delta_2^2$ by Proposition \ref{cor2}, and the third fraction almost surely converges to 1 by (i). The overall analysis shows that 
	\begin{equation}
		\limsup_{j\rightarrow\infty}\frac{S_2(T_j)}{\log(T_j)}\leq\frac{2}{\Delta_2^2}\label{mm3}
	\end{equation}
	with probability 1. In addition, since $\Prob(A_{T_k+1}=2|\Hp_{T_{k}})\Big\lfloor\frac{1}{\Prob(A_{T_k+1}=2|\Hp_{T_{k}})}\Big\rfloor$ almost surely converges to 1 by (i), we know that $\lim_{j\rightarrow\infty}\frac{S_2(T_j)}{j}=1$ with probability 1 by Stolz-Ces\`aro Theorem \cite{nagystolz}. Given that $B(T_j)=j$, \eqref{mm3} leads to
	\begin{equation}
		\limsup_{j\rightarrow\infty}\frac{B(T_j)}{\log(T_j)}\leq\frac{2}{\Delta_2^2}\label{mm4}
	\end{equation}
	almost surely. With \eqref{mm3} and \eqref{mm4}, proof is almost complete.

	 Given any $0<\delta<1$, we have $S_2(T_k)\leq \frac{2}{\delta\Delta_2^2}\log(T_k)$ almost surely for any big enough $k$ by \eqref{mm3}. Now if $T_k< T\leq T_{k+1}$, then 
	\begin{align*}
		&\frac{\delta\Delta_2^2}{2}S_2(T)=\frac{\delta\Delta_2^2}{2}\Big(\frac{T_{k+1}-T}{T_{k+1}-T_k}S_2(T_k)+\frac{T-T_k}{T_{k+1}-T_k}S_2(T_{k+1})\Big)\\
		&\leq \frac{T_{k+1}-T}{T_{k+1}-T_k}\log(T_k)+\frac{T-T_k}{T_{k+1}-T_k}\log(T_{k+1})\\
		&\leq \log\Big(\frac{T_{k+1}-T}{T_{k+1}-T_k}T_k+\frac{T-T_k}{T_{k+1}-T_k}T_{k+1}\Big)=\log(T)
	\end{align*}
	where the second inequality follows from the concavity of $\log$. Consequently, almost surely we have $\limsup_{T\rightarrow\infty}\frac{S_2(T)}{\log(T)}\leq \frac{2}{\delta\Delta_2^2}$, which leads to $\limsup_{T\rightarrow\infty}\frac{S_2(T)}{\log(T)}\leq \frac{2}{\Delta_2^2}$ due to $0<\delta<1$ being arbitrary. The last inequality, however, finishes the $R(T)$ bound part of Theorem \ref{thm6} with the help of (ii). As for $B(T)$, we know that $B(T)=B(T_j)$ if $T_j\leq T<T_{j+1}$ by its definition. Combining this observation with \eqref{mm4} leads to $B(T)=B(T_j)\leq(1+\delta)\log(T_j)\leq(1+\delta)\log(T)$ almost surely for big enough $T$ and some $\delta>0$. Since $\delta$ is arbitrary, we have $\limsup_{T\rightarrow\infty}\frac{B(T)}{\log(T)}\leq \frac{2}{\Delta_2^2}$ almost surely, and this finishes the proof of Theorem \ref{thm6} in the case of two-armed bandits.

\bibliography{sample}

\newpage
\onecolumn
\appendix

\subsection{OUTLINE}
The appendix is organized as follows.
\begin{enumerate}
	\item Section \ref{tools} states technical tools necessary for our proofs.
	\item In Section \ref{characteristics}, we analyze the asymptotic behavior of the sampling process of Thompson sampling. The proofs of the three important results stated in Section \ref{2armed} are provided in this section.
	\item Section \ref{proof1} and \ref{t2p} respectively provide the proof of Theorem \ref{thm5} and Theorem \ref{thm6}.
\end{enumerate}

\subsection{TECHNICAL TOOLS}\label{tools}
Here we recall a couple crucial results for our proofs.
\subsubsection{Gaussian Tail Bounds}
Let $X$ be a standard normal random variable, i.e. $X\sim\mathcal{N}(0,1)$, and $\Prob(X\geq \delta)=Q(\delta)$ for any $\delta\in\mathbb{R}$. Proposition 2.1.2 of \cite{vershynin2019high} shows that
\begin{equation}
	\Big(\frac{1}{\delta}-\frac{1}{\delta^3}\Big)\frac{\exp(-\frac{\delta^2}{2})}{\sqrt{2\pi}}\leq Q(\delta)\leq \frac{1}{\delta}\frac{\exp(-\frac{\delta^2}{2})}{\sqrt{2\pi}}\label{e1},
\end{equation}
if $\delta>0$. 
\subsubsection{Stolz-Ces\`aro Theorem}
Let $\{b_t\}_{t=1}^\infty$ is a sequence of positive real numbers such that $\sum_{t=1}^\infty b_t=\infty$, then for any real sequence $\{a_t\}_{t=1}^\infty$ we have
\begin{equation}
	\liminf_{t\rightarrow\infty}\frac{a_t}{b_t}\leq\liminf_{t\rightarrow\infty}\frac{\sum_{n=1}^t a_n}{\sum_{n=1}^t b_n}\leq\limsup_{t\rightarrow\infty}\frac{\sum_{n=1}^t a_n}{\sum_{n=1}^t b_n}\leq\limsup_{t\rightarrow\infty}\frac{a_t}{b_t}.\label{sct}
\end{equation}
See \cite{nagystolz} for the proof of this theorem.
\subsubsection{Related Convexity Properties}\label{cvx}
Suppose $f:\mathbb{R}\rightarrow\mathbb{R}$ and its second derivative $f''(x)>0$ for any $x\in\mathbb{R}$. If there exists $a$ such that $f'(a)=0$, then $f(a)\leq f(x)$ for any $x$ and $f(b)\leq f(\hat{x})$ for any $a\leq b\leq \hat{x}$ (see Section 3.1.3 of \cite{boyd2004convex}). In addition, for any $x\in[a,b]\subset\mathbb{R}$ we have
\begin{equation*}
	\max\{f(a),f(b)\}\geq f(x)
\end{equation*} 
by Proposition 4 of \cite{abeille2017linear}.

\subsubsection{Algorithm Independent Results}

The following results apply to any algorithm the agent employs for the setting described in Section \ref{bab}.

\begin{lemma}[Lemma 5 of \cite{russo2016simple}]\label{lem1}
	Let $\{Y_t\}$ be an i.i.d. sequence of real-valued random variables with finite variance and let $\{X_t\}$ be a sequence of binary random variables. Suppose each sequence is adapted to the filtration $\{\F_t\}$, and define $Z_n=\Prob(X_t=1|\F_{t-1})$. If, conditioned on $\F_{t-1}$, each $Y_t$ is independent of $X_t$, then we have,
	\begin{equation*}
		\lim_{T\rightarrow\infty}\frac{\sum_{t=1}^T X_tY_t}{\sum_{t=1}^{T}Z_t}=\Ex[Y_1]\quad\text{a.s. on}\quad\sum_{t=1}^{\infty}Z_t=\infty
	\end{equation*}
	and
	\begin{equation*}
		\sup_{T}|\sum_{t=1}^{T}X_tY_t|<\infty\quad\text{a.s. on}\quad\sum_{t=1}^{\infty}Z_t<\infty.
	\end{equation*}	
\end{lemma}
The following proposition is an immediate consequence of Lemma \ref{lem1}.
\begin{lemma}\label{cor1}
	Let the setup be as described in Section \ref{bab}, then regardless of any policy the agent follows we have for any $1\leq i,j\leq I$:
	\begin{equation}
		\lim_{T\rightarrow\infty}\frac{\sum_{t=1}^{T}\mathbbm{1}_{\{A_t=i\}}}{\sum_{t=1}^{T}\Prob(A_t=i|\Hp_{t-1})}=1\quad\text{a.s. on}\quad\sum_{t=1}^{\infty}\Prob(A_t=i|\Hp_{t-1})=\infty\label{e2},
	\end{equation}
	\begin{equation}
		\sum_{t=1}^{\infty}\mathbbm{1}_{\{A_t=i\}}<\infty\quad\text{a.s. on}\quad\sum_{t=1}^{\infty}\Prob(A_t=i|\Hp_{t-1})<\infty\label{e3},
	\end{equation}
	and
	\begin{equation}
		\lim_{T\rightarrow\infty}\frac{\sum_{t=1}^{T}\mathbbm{1}_{\{A_t=i\}}(Y_{j,t}-\mu_j)}{N_i(T)}=0\quad\text{a.s. on}\quad\sum_{t=1}^{\infty}\mathbbm{1}_{\{A_t=i\}}=\infty\label{e4}.
	\end{equation}
\end{lemma}

Notice that the combination of \eqref{e2} and \eqref{e3} leads to the following outcome:
\begin{equation}
	\sum_{t=1}^{\infty}\mathbbm{1}_{\{A_t=i\}}=\infty\quad\text{if and only if}\quad\sum_{t=1}^{\infty}\Prob(A_t=i|\Hp_{t-1})=\infty\label{e5}
\end{equation}
with probability 1. 

\begin{IEEEproof}[\textbf{Proof of Lemma \ref{cor1}}]
	If we let $X_t=\mathbbm{1}_{\{A_t=i\}}$, $Y_t=1$, and $\F_{t-1}=\Hp_{t-1}$, then equation \eqref{e2} and \eqref{e3} immediately follow from Lemma \ref{lem1}. 
	
	To achieve the final result \eqref{e4}, we now let $X_t=\mathbbm{1}_{\{A_t=i\}}$, $Y_t=Y_{j,t}-\mu_j$, and $\F_{t-1}=\{\Hp_t,Y_{j,1}-\mu_j,Y_{j,2}-\mu_j,...,Y_{j,t}-\mu_j\}$ with $\F_{0}=\emptyset$. Here both $\mathbbm{1}_{\{A_t=i\}}$ and $Y_{j,t}-\mu_j$ are adapted to $\F_t$. In addition, the probability of sampling the $i^{th}$ action depends only on the past observations $\Hp_{t-1}$, i.e.
	\begin{equation}
		\Prob(A_t=i|\Hp_{t-1})=\Prob(A_t=i|\F_{t-1}).\label{ee1}
	\end{equation} 
	This equality and the fact that $Y_{j,t}-\mu_j$ is independent of $\F_{t-1}$ mean that $X_t$, $Y_t$ and $\F_t$ satisfy the conditions of Lemma \ref{lem1}. As a result, we conclude: 
	\begin{equation*}
		\lim_{T\rightarrow\infty}\frac{\sum_{t=1}^{T}\mathbbm{1}_{\{A_t=i\}}(Y_{j,t}-\mu_j)}{\sum_{t=1}^{T}\Prob(A_t=i|\Hp_{t-1})
		}=0\quad\text{a.s. on}\quad\sum_{t=1}^{\infty}\Prob(A_t=i|\Hp_{t-1})=\infty.
	\end{equation*}
	Note that we again used \eqref{ee1}. This result proves \eqref{e4}, since $N_i(T)$ and $S_i(T)$ can be exchanged with the help of \eqref{e2} and \eqref{e5}.
\end{IEEEproof}

\subsection{ACTION SELECTION CHARACTERISTICS OF THOMPSON SAMPLING}\label{characteristics}
In this section, we provide results that describe how the Thompson sampling agent samples her actions in the setting of Section \ref{setting}. We prove Proposition \ref{cor2} at the end of this section.

\subsubsection{Basic Properties}
\begin{lemma}\label{lem2}
	For any $i$ and $x\in\mathbb{R}$, a.s. on $\sum_{t=1}^\infty\mathbbm{1}_{\{A_t=i\}}=\infty$ we have 
	\begin{equation*}
		\lim_{t\rightarrow\infty}\Prob(\theta_i(t)\geq x|\Hp_{T_{b(t)}})= 0,
	\end{equation*}
	if $x>\mu_i$, and
	\begin{equation*}
		\lim_{t\rightarrow\infty}\Prob(\theta_i(t)\geq x|\Hp_{T_{b(t)}})= 1
	\end{equation*}
	if $x<\mu_i$.
	\begin{IEEEproof}
		Given the sampling process for Thompson sampling, we have
		\begin{equation*}
			\Prob(\theta_i(t)\geq x|\Hp_{T_{b(t)}})=Q\Bigg(\frac{x-\hat{\mu}_i(T_{b(t)})}{\sqrt{\hat{\sigma}^2_i(T_{b(t)})}}\Bigg)=Q\Big(\sqrt{1+N_i(T_{b(t)})}(x-\hat{\mu}_i(T_{b(t)}))\Big).
		\end{equation*}
		On $\{\sum_{t=1}^\infty\mathbbm{1}_{\{A_t=i\}}=\infty\}$, $x-\hat{\mu}_i(T_{b(t)})$ almost surely converges to $x-\mu_i$ by Lemma \ref{cor1} and the fact that $T_{b(t)}\rightarrow\infty$ with probability 1. As a result, almost surely on $\sum_{t=1}^\infty\mathbbm{1}_{\{A_t=i\}}=\infty$, $\sqrt{1+N_i(T_{b(t)})}(x-\hat{\mu}_i(T_{b(t)})$ will either diverge to $\infty$ or $-\infty$ depending on $x-\mu_i$ being positive or negative respectively. This finishes the proof since $Q(.)$ will either converge to 0 or 1 depending on the sign of $x-\mu_i$.
	\end{IEEEproof}
\end{lemma}

\begin{proposition}\label{prop6}
	For any $i$, we have
	\begin{equation}
		\Prob\Big(\sum_{t=1}^{\infty}\mathbbm{1}_{\{A_t=i\}}=\infty\Big)=1,\label{ff1}
	\end{equation}
	and the optimal action is selected more frequently than the other arms:
	\begin{equation}
		\Prob\Big(\lim_{T\rightarrow\infty}\frac{\sum_{t=1}^{T}\mathbbm{1}_{\{A_t=1\}}}{T}=1\Big)=1.\label{ff2}
	\end{equation}
\end{proposition}

\begin{IEEEproof}
	We first prove \eqref{ff1}. Let $E_i=\{\sum_{t=1}^{\infty}\mathbbm{1}_{\{A_t=i\}}=\infty\}$ and pick $x>\mu_1$. Then
	\begin{align}
		\Prob(A_t=i|\Hp_{T_{b(t)}})&\geq\Prob(\theta_i(t)\geq x, \theta_j(t)<x\text{ for }j\neq i|\Hp_{T_{b(t)}})\nonumber\\
		&=\Prob(\theta_i(t)\geq x|\Hp_{T_{b(t)}})\prod_{j\neq i}\Prob(\theta_j(t)< x|\Hp_{T_{b(t)}})\label{e8}
	\end{align}
	where \eqref{e8} follows from the fact that conditioned on the past observations, $\theta_i$s are sampled from independent Gaussian distributions. By Lemma \ref{lem2}, we know $\Prob(\theta_j(t)< x|\Hp_{T_{b(t)}})$ converges to 1 almost surely on $E_j$, and it remains a positive constant after a finite amount passes on $E_j^c$. Recall that $\theta_j(t)$ is  a Gaussian random variable conditioned on $\Hp_{T_{b(t)}}$. Combining this observation with \eqref{e8} leads to
	\begin{equation*}
		\liminf_{t\rightarrow\infty}\Prob(A_t=i|\Hp_{T_{b(t)}})>0
	\end{equation*}
	almost surely on $E_i^c$. Consequently,
	\begin{equation*}
		\sum_{t=1}^{\infty}\Prob(A_t=i|\Hp_{T_{b(t)}})=\infty
	\end{equation*}
	almost surely on $E_i^c$. However, by \eqref{e5}, this final equality means that $\Prob(E_i^c)=0$, which proves \eqref{ff1}.
	
	To prove \eqref{ff2}, let  $\mu_j<x<\mu_1$ for all $j\neq 1$. Similar to the earlier analysis, we have
	\begin{align*}
		\Prob(A_t=1|\Hp_{T_{b(t)}})\geq\Prob(\theta_1(t)\geq x|\Hp_{T_{b(t)}})\prod_{j\neq 1}\Prob(\theta_j(t)< x|\Hp_{T_{b(t)}})
	\end{align*}
	Since every arm is pulled infinitely often with probability 1 by \eqref{ff1}, using Lemma \ref{lem2} leads to the right hand side of the preceding equation almost surely converging to 1. As a result,
	\begin{equation*}
		\lim_{t\rightarrow\infty}\Prob(A_t=1|\Hp_{T_{b(t)}})=1
	\end{equation*} 
	almost surely. This equality leads to 
	\begin{equation*}
		\lim_{T\rightarrow\infty}\frac{\sum_{t=1}^T \Prob(A_t=1|\Hp_{T_{b(t)}})}{T}=1
	\end{equation*}
	almost surely; consequently,
	\begin{equation*}
		\lim_{T\rightarrow\infty}\frac{\sum_{t=1}^{T}\mathbbm{1}_{\{A_t=1\}}}{T}=1
	\end{equation*}
	almost surely by Lemma \ref{cor1}.
\end{IEEEproof}
\subsubsection{The Relationship Between $N_i(T)$, $S_i(T)$, and $R_i(T)$}
Note that Proposition \ref{prop6} and \eqref{e5} leads to
\begin{equation}
	\Prob\Big(\sum_{t=1}^{\infty}\Prob(A_t=i|\Hp_{t-1})=\infty\Big)=1,\label{jjjjj1}
\end{equation}
and with the help of Lemma \ref{cor1} we have
\begin{equation*}
	\lim_{T\rightarrow\infty}\frac{N_i(T)}{S_i(T)}=1
\end{equation*}
with probability 1. In addition, the combination of Lemma \ref{cor1} and Proposition \ref{prop6} proves that
\begin{equation*}
	\lim_{T\rightarrow\infty}\frac{\sum_{t=1}^{T}\mathbbm{1}_{\{A_t=i\}}(Y_{1,t}-\mu_1+\mu_i-Y_{i,t})}{N_i(T)}= 0
\end{equation*}
almost surely for any $i\geq 2$, and this limit implies the following 
\begin{equation*}
	\lim_{T\rightarrow\infty}\frac{R_i(T)}{N_i(T)}=\Delta_i
\end{equation*} 
with probability 1 given the definition of $R_i(T)$. Ultimately the preceding discussion leads to
\begin{equation}
	\lim_{T\rightarrow\infty}\frac{R_i(T)}{S_i(T)}=\Delta_i\label{f2}
\end{equation}
almost surely for any suboptimal arm $i$.

\subsubsection{Proof of Proposition \ref{cor2}}

To prove Proposition \ref{cor2}, we will prove an upper bound and a lower bound on the probability of selecting the suboptimal arm $i$. We start with the lower bound.

\begin{proposition}\label{prop3}
	For any $i\geq 2$ and $\delta>0$, there exist almost surely finite random $K_N$ and $K_S$ such that if $k\geq K_N$
	\begin{equation}
		\exp(-\frac{1}{2}(1+\delta)\Delta_i^2N_i(T_{k}))\leq\Prob(A_{T_k+1}=i|\Hp_{T_{k}})\label{e10}
	\end{equation}
	and if $k\geq K_S$
	\begin{equation}
		\exp(-\frac{1}{2}(1+\delta)\Delta_i^2S_i(T_{k}))\leq\Prob(A_{T_k+1}=i|\Hp_{T_{k}})\label{e11}
	\end{equation}
	\begin{IEEEproof}
		Firstly pick any $\epsilon>0$, then similar to the proof of Proposition \ref{prop6} we have
		\begin{align*}
			\Prob(A_{T_k+1}=i|\Hp_{T_{k}})\geq\Prob(\theta_i(T_k+1)\geq \mu_1+\epsilon|\Hp_{T_{k}})\prod_{l\neq i}\Prob(\theta_l(T_k+1)< \mu_1+\epsilon|\Hp_{T_{k}}).
		\end{align*}
		We know that 
		\begin{equation*}
			\Prob(\theta_l(T_j+1)< \mu_1+\epsilon|\Hp_{T_{j}})=1-Q\Big((\mu_1+\epsilon-\hat{\mu}_l(T_{j}))\sqrt{1+N_i(T_{j})}\Big).
		\end{equation*}
		By Proposition \ref{prop6} and Lemma \ref{cor1}, $\hat{\mu}_l(T_{j})$ almost surely converges to $\mu_l$ as $j\rightarrow\infty$. Consequently, $\Prob(\theta_l(T_j+1)< \mu_1+\epsilon|\Hp_{T_{j}})$ converges to $1$ with probability $1$ as $j\rightarrow\infty$. Similarly for $\Prob(\theta_i(T_j+1)\geq \mu_1+\epsilon|\Hp_{T_{j}})$, we have
		\begin{equation*}
			\Prob(\theta_i(T_j+1)\geq \mu_1+\epsilon|\Hp_{T_{j}})=Q\Big((\mu_1+\epsilon-\hat{\mu}_i(T_{j}))\sqrt{1+N_i(T_{j})}\Big),
		\end{equation*}
		and $\mu_1+\epsilon-\hat{\mu}_i(T_{b(t)})$ almost surely converges $\Delta_i+\epsilon$ by Lemma \ref{cor1} and Proposition \ref{prop6}. These convergence results and the monotonically decreasing nature of $Q(.)$ mean that there exists an almost surely finite random $K$ such that if $k\geq K$
		\begin{align*}
			&\Prob(A_{T_k+1}=i|\Hp_{T_{k}})\geq\frac{Q\Big((\Delta_i+2\epsilon)\sqrt{1+N_i(T_{k})}\Big)}{2}\\
			&\geq \frac{1}{2\sqrt{2\pi}}\Bigg(\frac{1}{(\Delta_i+2\epsilon)\sqrt{1+N_i(T_{k})}}-\frac{1}{((\Delta_i+2\epsilon)\sqrt{1+N_i(T_{k})})^3}\Bigg)\exp(-\frac{1}{2}(\Delta_i+2\epsilon)^2(1+N_i(T_{k}))), 
		\end{align*}
		where the last step follows from \eqref{e1}. Since the exponential functions dominate the polynomials for large values and $N_i(T_{k})$ diverges to infinity almost surely, we know that there exists an almost surely finite random $K_N$ such that if $k\geq K_N$
		\begin{equation*}
			\Prob(A_{T_k+1}=i|\Hp_{T_{k}})\geq \exp(-\frac{1}{2}(\Delta_i+3\epsilon)^2N_i(T_{k})).
		\end{equation*}
		Letting $\epsilon=\frac{\sqrt{1+\delta}\Delta_i-\Delta_i}{3}$ proves \eqref{e10}.
		
		To prove \eqref{e11}, note that for any $\gamma>0$ there exists an almost surely finite random $L$ such that if $k\geq L$
		\begin{equation*}
			(1+\gamma)S_i(T_{k})\geq N_i(T_{k})
		\end{equation*}
		by Lemma \ref{cor1} and \eqref{jjjjj1}. If we let $K_S=\max\{K_N,L\}$ with the appropriate $\delta$ and $\gamma$ values, we achieve \eqref{e11}.
	\end{IEEEproof}
\end{proposition}
We now prove the upper bound.

\begin{proposition}\label{prop4}
	For any $i\geq 2$ and $1>\delta>0$, there exist almost surely finite random $K_N$ and $K_S$ such that if $k\geq K_N$
	\begin{equation}
		\exp(-\frac{1}{2}(1-\delta)\Delta_i^2N_i(T_{k}))\geq\Prob(A_{T_k+1}=i|\Hp_{T_{k}})\label{e12}
	\end{equation}
	and if $k\geq K_S$
	\begin{equation}
		\exp(-\frac{1}{2}(1-\delta)\Delta_i^2S_i(T_{k}))\geq\Prob(A_{T_k+1}=i|\Hp_{T_{k}})\label{e13}
	\end{equation}
	
	\begin{IEEEproof}
		Firstly,
		\begin{align*}
			\Prob(A_{T_j+1}=i|\Hp_{T_{j}})&\leq\Prob(\theta_i(T_j+1)\geq\theta_1(T_j+1)|\Hp_{T_{j}})\\
			&=Q\Big(\frac{\hat{\mu}_1(T_{j})-\hat{\mu}_i(T_{j})}{\sqrt{\hat{\sigma}_1^2(T_{j})+\hat{\sigma}_i^2(T_{j})}}\Big)
		\end{align*}
		By Lemma \ref{cor1} and Proposition \ref{prop6}, we know that $\hat{\mu}_1(T_{j})-\hat{\mu}_i(T_{j})$ almost surely converges to $\Delta_i$. Also, $\frac{\hat{\sigma}_1^2(T_{j})}{\hat{\sigma}_i^2(T_{j})}=\frac{1+N_i(T_{j})}{1+N_1(T_{j})}$ almost surely converges to 0 by Proposition \ref{prop6}. These results mean that for any $\gamma>0$ there exists an almost surely finite random $K\in\mathbb{Z}^+$ such that if $k\geq K$
		\begin{equation*}
			\frac{\hat{\mu}_1(T_{k})-\hat{\mu}_i(T_{k})}{\sqrt{\hat{\sigma}_1^2(T_{k})+\hat{\sigma}_i^2(T_{k})}}\geq \frac{\Delta_i}{\sqrt{(1+\gamma)\hat{\sigma}_i^2(T_{k})}}=\frac{1}{\sqrt{1+\gamma}}\sqrt{(1+N_i(T_{k}))}\Delta_i
		\end{equation*}
		Then by \eqref{e1}, there exists an almost surely finite random $K_N$ such that if $k\geq K_N$
		\begin{equation*}
			\Prob(A_{T_k+1}=i|\Hp_{T_{k}})\leq \exp(-\frac{1}{2}\frac{\Delta_i^2}{1+\gamma}N_i(T_{k})).
		\end{equation*}
		Note that we ignored $\frac{\sqrt{1+\gamma}}{2\pi\Delta_i\sqrt{1+N_i(T_{k})}}$ and the extra $1$ in the exponent since $N_i(T_{k})$ almost surely diverges by Proposition \ref{prop6} so the upper bound can be adjusted by appropriately changing $\gamma$. Finally, letting $\gamma=\frac{1}{1-\delta}-1$ proves \eqref{e12}.
		
		The proof of \eqref{e13} is analogous to the proof of \eqref{e11} in Proposition \ref{prop3}.
	\end{IEEEproof}
\end{proposition}

Proposition \ref{cor2} is an immediate result of Proposition \ref{prop3} and \ref{prop4}, since those results are applicable for any rational $0<\delta<1$. 

\subsection{PROOF OF THEOREM \ref{thm5}}\label{proof1}

We start with the lower bound
\begin{equation*}
	\Prob\Bigg(\liminf_{T\rightarrow\infty} \frac{R(T)}{\log(T)}\geq\sum_{i=2}^I\frac{2}{\Delta_i}\Bigg)=1
\end{equation*}	
in the next section and later provide the upper bound
\begin{equation*}
	\Prob\Bigg(\limsup_{T\rightarrow\infty} \frac{R(T)}{\log(T)}\leq\sum_{i=2}^I\frac{2}{\Delta_i}\Bigg)=1.
\end{equation*}
The combination of these results prove Theorem \ref{thm5}.

\subsubsection{Lower Bound}\label{lower}
We first present the following important proposition.
\begin{proposition}\label{thm1}
	For any batched bandit setting described in Section \ref{setting}, the total measurement effort Thompson sampling allocates to the suboptimal arm $i$ follows the next equation:
	\begin{equation*}
		\liminf_{j\rightarrow\infty} \frac{S_i(T_{j})}{\log(\frac{\Delta_i^2}{2}T_{j})}\geq \frac{2}{\Delta_i^2}
	\end{equation*}
	almost surely for any suboptimal arm $i$.
\end{proposition}
\begin{IEEEproof}
	Fix $\delta>0$ and let $U(\delta)=\{S_i(T_j)\geq \frac{2}{(1+\delta)\Delta_i^2}\log(\frac{\Delta_i^2}{2}T_j) \text{ for infinitely many } j\}$, $L(\delta)=\{S_i(T_j)< \frac{2}{(1+\delta)\Delta_i^2}\log(\frac{\Delta_i^2}{2}T_j) \text{ for infinitely many } j\}$. Then by Proposition \ref{cor2}, there exists an almost surely finite random $K$ such that if $k\geq K$, then
	\begin{equation*}
		\exp(-\frac{1}{2}(1+\delta)\Delta_i^2S_i(T_k))\leq\Prob(A_{T_k+1}=i|\Hp_{T_{k}}).
	\end{equation*}
	This inequality leads to the following by \eqref{sit1}
	\begin{equation}
		\frac{\Delta_i^2}{2}S_i(T_{k})+\frac{\Delta_i^2}{2}(T_{k+1}-T_k)\exp(-\frac{1}{2}(1+\delta)\Delta_i^2S_i(T_k))\leq \frac{\Delta_i^2}{2}S_i(T_{k+1}).\label{k1}
	\end{equation}
	The left hand side of the preceding equation is strictly convex in $\frac{\Delta_i^2}{2}S_i(T_{k})$, since its second derivative with respect to $\frac{\Delta_i^2}{2}S_i(T_{k})$ is positive everywhere on $\mathbb{R}$. This convex function achieves its minimum value at $\frac{\Delta_i^2}{2}S_i(T_{k})=\frac{\log(\frac{\Delta_i^2}{2}(T_{k+1}-T_k)(1+\delta))}{1+\delta}$, i.e. the point at which the first derivative is 0; consequently, this minimal point leads to
	\begin{equation}
		\frac{\Delta_i^2}{2}S_i(T_{k+1})\geq\frac{\log(\frac{\Delta_i^2}{2}(T_{k+1}-T_k)(1+\delta))}{1+\delta}+\frac{1}{1+\delta}=\frac{\log(e\frac{\Delta_i^2}{2}(T_{k+1}-T_k)(1+\delta))}{1+\delta}\label{h1}
	\end{equation}
	almost surely for any $k\geq K$. This lower bound means that if $\frac{\Delta_i^2}{2}S_i(T_{k})\geq \frac{1}{1+\delta}\log(\frac{\Delta_i^2}{2}T_k)$, there are two options. Firstly, if $\frac{1}{1+\delta}\log(\frac{\Delta_i^2}{2}T_k)\geq\frac{1}{1+\delta} \log(\frac{\Delta_i^2}{2}(T_{k+1}-T_k)(1+\delta))$, then the left hand side of $\eqref{k1}$ achieves its minimum at $\frac{\Delta_i^2}{2}S_i(T_{k})=\frac{\log(\frac{\Delta_i^2}{2}T_k)}{1+\delta}$ on $[\frac{\log(\frac{\Delta_i^2}{2}T_k)}{1+\delta},\infty)$, i.e. the boundary point closest to the global minimum (see Section \ref{cvx}), with the minimum being $\frac{1}{1+\delta}\log(\frac{\Delta_i^2}{2}T_k)+\frac{T_{k+1}-T_k}{T_k}$. Since $\log(1+x)\leq x$ for any $x>-1$, we have
	\begin{equation*}
		\frac{T_{k+1}-T_k}{(1+\delta)T_k}\geq \frac{1}{1+\delta}\log(1+\frac{T_{k+1}-T_k}{T_k})=\frac{1}{1+\delta}\log(\frac{T_{k+1}}{T_k}).
	\end{equation*}
	This analysis shows that $\frac{\Delta_i^2}{2}S_i(T_{k+1})\geq\frac{1}{1+\delta}\log(\frac{\Delta_i^2}{2}T_{k+1})$ almost surely for any $k\geq K$, if $\frac{\Delta_i^2}{2}S_i(T_{k})\geq\frac{1}{1+\delta}\log(\frac{\Delta_i^2}{2}T_k)$ and $\frac{1}{1+\delta}\log(\frac{\Delta_i^2}{2}T_k)\geq\frac{1}{1+\delta} \log(\frac{\Delta_i^2}{2}(T_{k+1}-T_k)(1+\delta))$. Now suppose $\frac{1}{1+\delta}\log(\frac{\Delta_i^2}{2}T_k)<\frac{1}{1+\delta} \log(\frac{\Delta_i^2}{2}(T_{k+1}-T_k)(1+\delta))$, then we have
	\begin{equation*}
		T_{k+1}=(T_{k+1}-T_k)+T_k<2(T_{k+1}-T_k)(1+\delta).
	\end{equation*} 
	This equation leads to $\frac{\Delta_i^2}{2}S_i(T_{k+1})\geq\frac{1}{1+\delta}\log(\frac{\Delta_i^2}{2}T_{k+1})$ almost surely for any $k\geq K$ by \eqref{h1}, if $\frac{1}{1+\delta}\log(\frac{\Delta_i^2}{2}T_k)<\frac{1}{1+\delta} \log(\frac{\Delta_i^2}{2}(T_{k+1}-T_k)(1+\delta))$. This overall analysis shows that almost surely $\frac{\Delta_i^2}{2}S_i(T_{k+1})\geq\frac{1}{1+\delta}\log(\frac{\Delta_i^2}{2}T_{k+1})$ for any $k\geq K$, if $\frac{\Delta_i^2}{2}S_i(T_{k})\geq\frac{1}{1+\delta}\log(\frac{\Delta_i^2}{2}T_{k})$. Consequently, this result proves $\Prob(U(\delta)\cap L(\delta))=0$.
	
	For the final part, by Proposition \ref{cor2}, we note that almost surely there exists a random $K\in\mathbb{Z}^+$ such that if $k\geq K$, then
	\begin{equation*}
		\exp(-\frac{1}{2}(1+\delta/2)\Delta_i^2S_i(T_k))\leq\Prob(A_{T_k+1}=i|\Hp_{T_{k}}).
	\end{equation*} 
	This lower bound mean that almost surely on $U(\delta)^c$:
	\begin{equation*}
		\Big(\frac{2}{\Delta_i^2}\Big)^{(\frac{1+\delta/2}{1+\delta})}\leq\liminf_{j\rightarrow\infty}\frac{\Prob(A_{T_j+1}=i|\Hp_{T_{j}})}{T_j^{-(\frac{1+\delta/2}{1+\delta})}}\leq\liminf_{j\rightarrow\infty}\frac{S_i(T_j)}{T_1+\sum_{m=2}^{j}\frac{T_m-T_{m-1}}{T_{m-1}^{(\frac{1+\delta/2}{1+\delta})}}}\leq \liminf_{j\rightarrow\infty}\frac{S_i(T_j)}{\sum_{t=1}^{T_j}\frac{1}{t^{(\frac{1+\delta/2}{1+\delta})}}}
	\end{equation*}
	where the second inequality follows from \eqref{maineq} and \eqref{sct}. The final inequality is the result of the batch nature of the bandit problem. In addition, $\sum_{t=1}^{T_j}\frac{1}{t^{(\frac{1+\delta/2}{1+\delta})}}$ scales with $T_j^{(\frac{\delta/2}{1+\delta})}$ as $j\rightarrow\infty$. However, this is a contradiction since on $U(\delta)^c$:
	\begin{equation*}
		\limsup_{j\rightarrow\infty} \frac{S_i(T_j)}{\log(\frac{\Delta_i^2}{2}T_j)}\leq \frac{2}{(1+\delta)\Delta_i^2}.
	\end{equation*}
	As a result, we see that $\Prob(U(\delta)^c)=0$.
	
	Finally, the overall analysis shows that $\Prob(U(\delta)\cap L(\delta)^c)=1$, which results in
	\begin{equation*}
		\liminf_{j\rightarrow\infty} \frac{S_i(T_j)}{\log(\frac{\Delta_i^2}{2}T_j)}\geq \frac{2}{(1+\delta)\Delta_i^2}
	\end{equation*}
	almost surely. This lower bound finishes the proof, since it is true for any rational $\delta$ in $(0,\infty)$.

\end{IEEEproof}

Now for any $T\in\mathbb{Z}^+$ there exists $j\in\mathbb{Z}^+$ such that $ T_{j-1}<T\leq T_j$. As a result, $S_i(T)\geq S_i(T_{j-1})$ and $\log(T)\leq\log(T_{j})$, which lead to
\begin{equation*}
	\frac{S_i(T)}{\log(\frac{\Delta_i^2}{2}T)}\geq \frac{S_i(T_{j-1})}{\log(\frac{\Delta_i^2}{2}T_{j})}=\frac{\log(\frac{\Delta_i^2}{2}T_{j-1})}{\log(\frac{\Delta_i^2}{2}T_{j})}\frac{S_i(T_{j-1})}{\log(\frac{\Delta_i^2}{2}T_{j-1})}.
\end{equation*}
Note that $S_i$ is a monotonically increasing function of $T$.
Since there are infinitely many batches with probability 1, as $T\rightarrow\infty$, $j$ that satisfies the $ T_{j-1}<T\leq T_j$ condition diverges to infinity almost surely. By combining this observation with the earlier inequality, we have
\begin{align*}
	\liminf_{T\rightarrow\infty}\frac{S_i(T)}{\log(\frac{\Delta_i^2}{2}T)}&\geq\liminf_{j\rightarrow\infty}\frac{\log(\frac{\Delta_i^2}{2}T_{j-1})}{\log(\frac{\Delta_i^2}{2}T_{j})}\frac{S_i(T_{j-1})}{\log(\frac{\Delta_i^2}{2}T_{j-1})}\\
	&\geq\frac{2}{\Delta_i^2}\liminf_{j\rightarrow\infty}\frac{\log(\frac{\Delta_i^2}{2}T_{j-1})}{\log(\frac{\Delta_i^2}{2}T_{j})}
\end{align*}
almost surely. The last inequality follows from Proposition \ref{thm1}.	This analysis and \eqref{f2} lead to
\begin{align}
	\liminf_{T\rightarrow\infty} \frac{R(T)}{\log(T)}&=\liminf_{T\rightarrow\infty}\sum_{i=2}^I\frac{R_i(T)}{S_i(T)}\frac{S_i(T)}{\log(T)}\nonumber\\
	&\geq\sum_{i=2}^I\frac{2}{\Delta_i}\liminf_{j\rightarrow\infty}\frac{\log(T_j)}{\log(T_{j+1})}\label{llll1}
\end{align}
Note that the $\log(\frac{\Delta_i^2}{2})$ term is ignored since $T$ and $T_j$s will almost surely diverge. However, by \eqref{ggg3} we know there exists an almost surely finite random $K$ such that if $k\geq K$ then $2T_k\geq T_{k+1}$, which leads to 
\begin{equation*}
	\frac{\log(T_k)}{\log(T_{k+1})}\geq\frac{\log(T_k)}{\log(2)+\log(T_k)}.
\end{equation*}
Considering this inequality, \eqref{llll1} leads to
\begin{equation*}
	\liminf_{T\rightarrow\infty} \frac{R(T)}{\log(T)}\geq\sum_{i=2}^I\frac{2}{\Delta_i}
\end{equation*}	
with probability 1.

\subsubsection{Upper Bound}\label{upper}

We first upper bound the total measurement effort $S_i(T_j)$.
\begin{proposition}\label{thm2}
	Suppose that with probability 1
	\begin{equation}
		\limsup_{j\rightarrow\infty}\log_{T_j}(T_{j+1}-T_j)<1,\label{k3}
	\end{equation}
	then 
	\begin{equation}
		\limsup_{j\rightarrow\infty}\frac{S_i(T_j)}{\log(\frac{\Delta_i^2}{2}T_j)}\leq \frac{2}{\Delta_i^2}
	\end{equation}
	almost surely for any suboptimal arm $i$.
	\begin{IEEEproof}
		First let $0<\delta<1$ and $E(\delta)=\{S_i(T_j)\leq \frac{2}{\delta\Delta_i^2}\log(\frac{\Delta_i^2}{2}T_j)\text{ for infinitely many }j\}$. On $E(\delta)^c$: 
		\begin{align*}
			\liminf_{j\rightarrow\infty}\frac{-\log(\Prob(A_{T_j+1}=i|\Hp_{T_{j}}))}{\log(\frac{\Delta_i^2}{2}T_j)}
			&=\liminf_{j\rightarrow\infty}\frac{-\log(\Prob(A_{T_j+1}=i|\Hp_{T_{j}}))}{S_i(T_j)}\frac{S_i(T_j)}{\log(\frac{\Delta_i^2}{2}T_j)}\\
			&\geq \frac{1}{\delta}
		\end{align*}
		almost surely by Proposition \ref{cor2}. This analysis leads to the following convergence result almost surely on $E(\delta)^c$:
		\begin{equation}
			\lim_{j\rightarrow\infty}\frac{\Prob(A_{T_j+1}=i|\Hp_{T_{j}})}{T_j^{-\frac{1+\delta}{2\delta}}}=0.\label{k4}
		\end{equation}
		By \eqref{k3}, there exists an almost surely finite random $C$ such that $CT_j\geq T_{j+1}$ for any $j$. Consequently, this upper bound leads to the following result almost surely on $E(\delta)^c$: 
		\begin{align*}
			\infty&>\limsup_{j\rightarrow\infty}\frac{S_i(T_j)}{T_1+\sum_{m=2}^{j}\frac{T_m-T_{m-1}}{T_{m-1}^{\frac{1+\delta}{2\delta}}}}\\
			&\geq \limsup_{j\rightarrow\infty}\frac{S_i(T_j)}{T_1+C^{\frac{1+\delta}{2\delta}}\sum_{t=T_1+1}^{T_j}\frac{1}{t^{\frac{1+\delta}{2\delta}}}}
		\end{align*}
		where the first inequality follows from \eqref{maineq} and \eqref{k4}. Since $\lim_{j\rightarrow\infty}\sum_{t=T_1+1}^{T_j}\frac{1}{t^{\frac{1+\delta}{2\delta}}}<\infty$, $\sup_{j}S_i(T_j)<\infty$ almost surely on $E(\delta)^c$. From here, we see that $\Prob(E(\delta)^c)=0$.
		
		Now we will analyze the sample paths on $E(\delta)\cap Q(\epsilon)$, where $Q(\epsilon)=\{T_{j+1}-T_j<T_j^\epsilon \text{ for all big enough } j\}$ for some $0<\epsilon<1$. Let $\gamma=\frac{1+\max\{\delta,\sqrt{\epsilon}\}}{2}$, then by Propositions \ref{cor2} and \ref{thm1} there exists an almost surely finite random $K$ such that if $k\geq K$:
		\begin{align}
			\Prob(A_{T_k+1}=i|\Hp_{T_{k}})&\leq\exp(-\frac{\gamma}{2}\Delta_i^2S_i(T_k))\label{k5}\\
			\gamma\log(\frac{\Delta_i^2}{2}T_k)&\leq\frac{\Delta_i^2}{2}S_i(T_k).\label{k6}
		\end{align} 
		Equation \eqref{k5} leads to the following almost surely if $k\geq K$:
		\begin{equation}
			\frac{\Delta_i^2}{2}S_i(T_{k+1})\leq \frac{\Delta_i^2}{2}S_i(T_{k})+(T_{k+1}-T_k)\frac{\Delta_i^2}{2}\exp(-\frac{\gamma}{2}\Delta_i^2S_i(T_k)).\label{k7} 
		\end{equation}
		Clearly \eqref{k7} is strictly convex in $\frac{\Delta_i^2}{2}S_i(T_k)$; as a result, \eqref{k7} achieves its maximum either on $\frac{\Delta_i^2}{2}S_i(T_k)=\gamma\log(\frac{\Delta_i^2}{2}T_k)$ or $\frac{\Delta_i^2}{2}S_i(T_k)=\frac{1}{\delta}\log(\frac{\Delta_i^2}{2}T_k)$ if $\gamma\log(\frac{\Delta_i^2}{2}T_k)\leq \frac{\Delta_i^2}{2}S_i(T_k)\leq\frac{1}{\delta}\log(\frac{\Delta_i^2}{2}T_k)$ (see Section \ref{cvx}). So, we will provide upper bounds depending on these boundary points. Firstly, if $\frac{\Delta_i^2}{2}S_i(T_k)=\frac{1}{\delta}\log(\frac{\Delta_i^2}{2}T_k)$, then
		\begin{equation}
			\frac{\Delta_i^2}{2}S_i(T_{k+1})\leq \frac{1}{\delta}\log(\frac{\Delta_i^2}{2}T_k)+\frac{T_{k+1}-T_k}{T_k^{\frac{\gamma}{\delta}}}\Big(\frac{\Delta_i^2}{2}\Big)^{1-\frac{\gamma}{\delta}}.
		\end{equation}
		Since $\frac{x}{1+x}\leq\log(1+x)$ for any $x>-1$ and $\gamma>\delta$, for all big enough $k$ we have
		\begin{align*}
			\frac{T_{k+1}-T_k}{T_k^{\frac{\gamma}{\delta}}}\Big(\frac{\Delta_i^2}{2}\Big)^{1-\frac{\gamma}{\delta}}&\leq \frac{1}{\delta} \frac{T_{k+1}-T_k}{T_{k+1}}\\
			&\leq \frac{1}{\delta}\log(1+\frac{T_{k+1}-T_k}{T_k})\\
			&=\frac{1}{\delta}\log(\frac{T_{k+1}}{T_k}).
		\end{align*}
		where the first inequality follows the fact that we are considering the sample paths on $Q(\epsilon)$. This analysis means that almost surely $\frac{\Delta_i^2}{2}S_i(T_{k+1})\leq\frac{1}{\delta}\log(T_{k+1})$ if $\frac{\Delta_i^2}{2}S_i(T_k)=\frac{1}{\delta}\log(\frac{\Delta_i^2}{2}T_k)$ for all big enough $k$ on $Q(\epsilon)$.
		
		Now suppose $\frac{\Delta_i^2}{2}S_i(T_k)=\gamma\log(\frac{\Delta_i^2}{2}T_k)$, then by \eqref{k7} we have the following inequality for any big enough $k\geq K$:
		\begin{align*}
			\frac{\Delta_i^2}{2}S_i(T_{k+1})&\leq \gamma\log(\frac{\Delta_i^2}{2}T_k)+\frac{T_{k+1}-T_k}{T_k^{\gamma^2}}(\frac{\Delta_i^2}{2})^{1-\gamma^2}\\
			&\leq \gamma\log(\frac{\Delta_i^2}{2}T_{k+1})+(\frac{\Delta_i^2}{2})^{1-\gamma^2}\\
			&\leq \frac{1}{\delta}\log(\frac{\Delta_i^2}{2}T_{k+1})
		\end{align*}
		where the second inequality follows from the fact that the sample path is on $Q(\epsilon)$ and $\gamma^2\geq\epsilon$. The final inequality is the result of $\gamma<1<1/\delta$.
		
		This overall analysis show that if $\gamma\log(\frac{\Delta_i^2}{2}T_k)\leq \frac{\Delta_i^2}{2}S_i(T_k)\leq\frac{1}{\delta}\log(\frac{\Delta_i^2}{2}T_k)$ then $\frac{\Delta_i^2}{2}S_i(T_{k+1})\leq\frac{1}{\delta}\log(\frac{\Delta_i^2}{2}T_{k+1})$ almost surely for all big enough $k$ on $Q(\epsilon)$. Consequently, this result and \eqref{k6} shows that 
		\begin{equation*}
			\limsup_{j\rightarrow\infty}\frac{S_i(T_j)}{\log(\frac{\Delta_i^2}{2}T_j)}\leq \frac{2}{\delta\Delta_i^2}
		\end{equation*}
		almost surely on $E(\delta)\cap Q(\epsilon)$. Since we know $\Prob(\cup_{n=2}^\infty Q(1-\frac{1}{n}))=1$ by \eqref{k3}, the preceding convergence result holds true almost surely on $E(\delta)$. Since $\Prob(E(\delta))=1$ for any rational $0<\delta<1$ by the initial analysis, we finish the proof.
	\end{IEEEproof}	
	
\end{proposition}
By this proposition, we have the following upper bound given any $0<\delta<1$:
\begin{equation*}
	S_i(T_k)\leq \frac{2}{\delta\Delta_i^2}\log(\frac{\Delta_i^2}{2}T_k)
\end{equation*}
almost surely for any big enough $k$. Now if $T_k< T\leq T_{k+1}$, then 
\begin{align*}
	S_i(T)&=\frac{T_{k+1}-T}{T_{k+1}-T_k}S_i(T_k)+\frac{T-T_k}{T_{k+1}-T_k}S_i(T_{k+1})\\
	&\leq \frac{2}{\delta\Delta_i^2}\Big(\frac{T_{k+1}-T}{T_{k+1}-T_k}\log(\frac{\Delta_i^2}{2}T_k)+\frac{T-T_k}{T_{k+1}-T_k}\log(\frac{\Delta_i^2}{2}T_{k+1})\Big)\\
	&\leq \frac{2}{\delta\Delta_i^2}\log\Big(\frac{\Delta_i^2}{2}\frac{T_{k+1}-T}{T_{k+1}-T_k}T_k+\frac{\Delta_i^2}{2}\frac{T-T_k}{T_{k+1}-T_k}T_{k+1}\Big)\\
	&\leq \frac{2}{\delta\Delta_i^2}\log(\frac{\Delta_i^2}{2}T)
\end{align*}
where the second inequality follows from the concavity of $\log$. Consequently, almost surely we have
\begin{equation*}
	\limsup_{T\rightarrow\infty}\frac{S_i(T)}{\log(\frac{\Delta_i^2}{2}T)}\leq \frac{2}{\delta\Delta_i^2},
\end{equation*}
which leads to
\begin{equation}
	\limsup_{T\rightarrow\infty}\frac{S_i(T)}{\log(\frac{\Delta_i^2}{2}T)}\leq \frac{2}{\Delta_i^2}\label{k10}
\end{equation}
due to $0<\delta<1$ being arbitrary.

Similar to the analysis at the end of Section \ref{lower}, we have
\begin{align*}
	\limsup_{T\rightarrow\infty}\frac{R(T)}{\log(T)}&=\limsup_{T\rightarrow\infty}\sum_{i=2}^I \frac{R_i(T)}{\log(T)}\\
	&=\limsup_{T\rightarrow\infty}\sum_{i=2}^I \frac{R_i(T)}{S_i(T)}\frac{S_i(T)}{\log(T)}\\
	&\leq\sum_{i=2}^I \frac{2}{\Delta_i}
\end{align*}
almost surely. The last step follows from \eqref{f2} and Proposition \ref{thm2}.

\subsection{Proof of Theorem \ref{thm6}}\label{t2p}
We first prove the fact that iPASE satisfies the conditions set in Section \ref{setting}. It is clear that by \eqref{inv}, $T_j$'s are adapted to $H_{T_{j-1}}$ and $\Prob(T_1=I)=1$. Now suppose $\Prob(T_j<\infty)=1$ for some $j\geq 1$. Then by the definitions of $\hat{\mu}$ and $\hat{\sigma}^2$ in Section \ref{setting}, we know $\hat{\sigma}^2_l(T_{j})>0$ and $|\hat{\mu}_l(T_{j})|<\infty$ for all $1\leq l\leq I$ with probability 1. This observation implies $\Prob(A_{T_j+1}=i|\Hp_{T_{j}})>0$ almost surely since it is the  probability that out of the conditionally independent random variables with distributions $\{\mathcal{N}(\hat{\mu}_l(T_{j}),\hat{\sigma}^2_l(T_{j}))\}_{l=1}^I$ the one with the distribution $\mathcal{N}(\hat{\mu}_i(T_{j}),\hat{\sigma}^2_i(T_{j}))$ being the biggest. Combining this analysis with \eqref{inv} shows $T_{j+1}-T_j<\infty$ almost surely; as a result, $\Prob(T_{j+1}<\infty)=1$. Finally the induction hypothesis proves the fact that there are almost surely infinite number of batches with iPASE. So iPASE satisfies the policy conditions in Section \ref{setting}, and the setting related results also apply to it.

Now we summarize the main steps of our analysis. Firstly, we prove that the asymptotic rate of the measurement allocated to suboptimal arm $i$ times $\Delta_i^2$ has to be same for all $i$. Since all suboptimal arms satisfy Proposition \ref{cor2}, $\Delta_{i}^2S_{i}(T_j)$ of one suboptimal arm can not dominate others' in the limit as the limit presented in Proposition \ref{cor2} does not allow it. That means $\Delta_{i_1}^2S_{i_1}(T_j)$ terms for each suboptimal arm has to be in a small neighborhood of each other infinitely often. However, as $j\rightarrow\infty$ Proposition \ref{cor2} will provide tighter bounds and \eqref{inv} will limit the growth of the $\Delta_{i_1}^2S_{i_1}(T_j)$ terms. This will prove the first step. Then in the second step, the combination of this result with Proposition \ref{cor2} will show that we can discard the $\max$ operator from the definition of $\Prob_{2}(T_j)$ and describe it in terms of any arbitrary suboptimal arm $i$. Since $-\log(\lfloor\Prob_{2}(T_j)\rfloor)$ lower bounds the term $\log(T_{j+1})$, a simple analysis will lead to a $O(\log(T))$ bound on $S_i(T)$. Finally, considering the fact that 
\begin{equation*}
	\sum_{i=2}^I\Prob(A_{T_j+1}=i|\Hp_{T_{j}})\Big\lfloor\frac{1}{\Prob_{2}(T_j)}\Big\rfloor\geq 1
\end{equation*}
for big $j$ values, i.e. the increase in the total measurement effort for the suboptimal arms in the $(j+1)^{th}$ batch (left-hand side) is bigger than the increase in the batch count (right-hand side), the $O(\log(T))$ bound on $S_i(T)$ will lead to a $O(\log(T))$ bound on $B(T)$. $R(T)=O(\log(T))$ will similarly follow with the help of \eqref{f2}.

\begin{proposition}\label{prop5}
	For any two suboptimal arms $i_1$ and $i_2$, we have the following total measurement effort allocation ratio with iPASE:
	\begin{equation*}
		\lim_{j\rightarrow\infty}\frac{\Delta_{i_1}^2S_{i_1}(T_j)}{\Delta_{i_2}^2S_{i_2}(T_j)}=1
	\end{equation*}
	almost surely.
	\begin{IEEEproof}
		Let's fix any $0<\delta<1$ and define $U(\delta)=\{\Delta_{i_1}^2S_{i_1}(T_j)(1+\delta)\geq\Delta_{i_2}^2S_{i_2}(T_j)(1-\delta)\text{ for infinitely many j}\}$. Then on $U(\delta)^c$, there exists an almost surely finite $K$ such that if $k\geq K$, then $(1+\delta)\Delta_{i_1}^2S_{i_1}(T_k)<\Delta_{i_2}^2S_{i_2}(T_j)(1-\delta)$. In addition, by Proposition \ref{cor2} and the fact that $S_i$s almost surely diverge by Proposition \ref{thm1}, there exists an almost surely finite $\hat{K}$ such that if $k\geq\hat{K}$, then $\Delta_{i_1}^2\Prob(A_{T_k+1}=i_1|\Hp_{T_{k}})\geq\exp(-\frac{\Delta_{i_1}^2}{2}(1+\delta)S_{i_1}(T_k))$ and $\Delta_{i_2}^2\Prob(A_{T_k+1}=i_2|\Hp_{T_{k}})\leq\exp(-\frac{\Delta_{i_2}^2}{2}(1-\delta)S_{i_2}(T_k))$. By combining the previous analysis, we see that almost surely on $U(\delta)^c$, if $k\geq\max\{K,\hat{K}\}$, then $\Delta_{i_1}^2\Prob(A_{T_k+1}=i_1|\Hp_{T_{k}})\geq\Delta_{i_2}^2\Prob(A_{T_k+1}=i_2|\Hp_{T_{k}})$. Considering \eqref{maineq}, this inequality means that 
		\begin{equation*}
			\liminf_{j\rightarrow\infty}\frac{\Delta_{i_1}^2S_{i_1}(T_j)}{\Delta_{i_2}^2S_{i_2}(T_j)}\geq 1
		\end{equation*}  
		almost surely on $U(\delta)^c$. Given the definition of $U(\delta)$, this lower bound proves $\Prob(U(\delta)^c)=0$. Note that, in addition to this result, we also proved that $\Delta_{i_1}^2\Prob(A_{T_k+1}=i_1|\Hp_{T_{k}})\geq\Delta_{i_2}^2\Prob(A_{T_k+1}=i_2|\Hp_{T_{k}})$ almost surely for all big enough $k$, if $(1+\delta)\Delta_{i_1}^2S_{i_1}(T_k)<\Delta_{i_2}^2S_{i_2}(T_j)(1-\delta)$. 
		
		For the last part of the proof, we now show that there exists an almost surely finite $K$ such that if $k\geq K$, then $(1+\delta)\Delta_{i_1}^2S_{i_1}(T_k)\geq\Delta_{i_2}^2S_{i_2}(T_j)(1-\delta)$. To prove this claim first note that the combination of Propositions \ref{cor2} and \ref{thm1} shows that $\Prob(A_{T_j+1}=1|\Hp_{T_{j}})\rightarrow 1$ almost surely. Then from the earlier analysis we know that almost surely for any big enough $k$, we have
		\begin{align}
			\Delta_{i_1}^2\Prob(A_{T_k+1}=i_1|\Hp_{T_{k}})\geq\Delta_{i_2}^2\Prob(A_{T_k+1}=i_2|\Hp_{T_{k}})\quad\text{ if }(1+\delta/2)\Delta_{i_1}^2S_{i_1}(T_k)<\Delta_{i_2}^2S_{i_2}(T_k)(1-\delta/2),\label{fe1}
		\end{align}
		and
		\begin{equation}
			S_{i}(T_{k+1})-S_{i}(T_k)=\Prob(A_{T_k+1}=i|\Hp_{T_{k}})\Big\lfloor\frac{1}{\Prob_{2}(T_k)}\Big\rfloor\leq 1.\label{fe2}
		\end{equation}
		These inequalities mean that there exists a random $\hat{K}\in\mathbb{Z}^+$ such that if $k\geq\hat{K}$ and $(1+\delta/2)\Delta_{i_1}^2S_{i_1}(T_k)\geq\Delta_{i_2}^2S_{i_2}(T_k)(1-\delta/2)$, then
		\begin{align}
			\frac{\Delta_{i_1}^2S_{i_1}(T_{k+1})}{\Delta_{i_2}^2S_{i_2}(T_{k+1})}&\geq\frac{\Delta_{i_1}^2S_{i_1}(T_{k})}{\Delta_{i_2}^2S_{i_2}(T_{k})+\Delta_{i_2}^2}\label{fe3}\\
			&\geq \frac{(1-\delta/2)\Delta_{i_1}^2S_{i_1}(T_{k})}{(1+\delta/2)\Delta_{i_1}^2S_{i_1}(T_{k})+(1-\delta/2)\Delta_{i_2}^2}\nonumber\\
			&\geq \frac{1-\delta}{1+\delta}\label{fe4}
		\end{align}
		Here, \eqref{fe3} follows from \eqref{fe2}. As for \eqref{fe4}, we know that $S_{i_1}(T_k)$ almost surely diverges, so $(1+\delta)\Delta_{i_1}^2S_{i_1}(T_{k})$ almost surely dominates the term $(1+\delta/2)\Delta_{i_1}^2S_{i_1}(T_{k})+(1-\delta/2)\Delta_{i_2}^2$. Also, in consideration of \eqref{fe1}, if $(1+\delta/2)\Delta_{i_1}^2S_{i_1}(T_{k+1})<\Delta_{i_2}^2S_{i_2}(T_{k+1})(1-\delta/2)$, then
		\begin{equation*}
			\frac{\Delta_{i_1}^2S_{i_1}(T_{k+1})}{\Delta_{i_2}^2S_{i_2}(T_{k+1})}\leq\frac{\Delta_{i_1}^2S_{i_1}(T_{k+2})}{\Delta_{i_2}^2S_{i_2}(T_{k+2})}.
		\end{equation*} 
		Due to the relationship described in \eqref{fe1}, the preceding monotonic behavior will continue at least until $(1+\delta/2)\Delta_{i_1}^2S_{i_1}(T_{k})\geq\Delta_{i_2}^2S_{i_2}(T_{k})(1-\delta/2)$ again. Coupling this observation with the result in \eqref{fe4} means that almost surely on $U(\delta/2)$ there exists a random $K\in\mathbb{Z}^+$ such that if $k\geq K$, then $(1+\delta)\Delta_{i_1}^2S_{i_1}(T_k)\geq\Delta_{i_2}^2S_{i_2}(T_j)(1-\delta)$. Since $\Prob(U(\delta/2))=1$, we have
		\begin{equation*}
			\liminf_{j\rightarrow\infty}\frac{\Delta_{i_1}^2S_{i_1}(T_{j})}{\Delta_{i_2}^2S_{i_2}(T_{j})}\geq\frac{1-\delta}{1+\delta}
		\end{equation*}
		almost surely for any rational $0<\delta<1$, which leads to
		\begin{equation*}
			\liminf_{j\rightarrow\infty}\frac{\Delta_{i_1}^2S_{i_1}(T_{j})}{\Delta_{i_2}^2S_{i_2}(T_{j})}\geq 1
		\end{equation*}
		with probability 1. This result finishes the proof since the indices $i_1$ and $i_2$ are arbitrary. 
		
	\end{IEEEproof}
\end{proposition} 
Now, using the preceding result, we will provide the rate at which each suboptimal arm is sampled with iPASE.

\begin{theorem}\label{thm3}
	For any suboptimal arm i, we have
	\begin{equation*}
		\limsup_{T\rightarrow\infty}\frac{S_i(T)}{\log(T)}\leq\frac{2}{\Delta_i^2}
	\end{equation*}
	almost surely with iPASE.
	\begin{IEEEproof}
		Firstly,
		\begin{equation*}
			\frac{S_i(T_j)}{\log(T_j)}\leq \frac{S_i(T_j)}{\log(\frac{1}{\Prob_{2}(T_{j-1})}-1)}
		\end{equation*}
		due to \eqref{inv}. In addition for any $0<\delta<1$, by Proposition \ref{cor2}, we know that there exists an almost surely finite $\hat{K}$ such that if $k\geq\hat{K}$, then
		\begin{equation*}
			\frac{1}{\Prob_{2}(T_{k-1})}-1\geq\min_{2\leq l\leq I}\exp((1-\delta/2)\frac{\Delta_l^2}{2}S_l(T_{k-1})).
		\end{equation*} 
		As a result of this equation and Proposition \ref{prop5}, there exists an almost surely finite $K\in\mathbb{Z}^+$ such that if $k\geq K$, then
		\begin{equation*}
			\frac{1}{\Prob_{2}(T_{k-1})}-1\geq\exp((1-\delta)\frac{\Delta_i^2}{2}S_i(T_{k-1})),
		\end{equation*}
		which in turn leads to
		\begin{equation*}
			\frac{S_i(T_k)}{\log(T_k)}\leq\frac{S_i(T_k)}{(1-\delta)\frac{\Delta_i^2}{2}S_i(T_{k-1})}
			\leq \frac{S_i(T_{k-1})+1}{(1-\delta)\frac{\Delta_i^2}{2}S_i(T_{k-1})}
		\end{equation*}
		where the rightmost inequality follows from the fact that \eqref{inv} limits the amount of measurement allocated for each suboptimal arm by $1$ for big $k$ values. Since $S_i$ almost surely diverges, and the preceding analysis applies to any rational $0<\delta<1$, we have
		\begin{equation*}
			\limsup_{j\rightarrow\infty}\frac{S_i(T_j)}{\log(T_j)}\leq\frac{2}{\Delta_i^2}
		\end{equation*}
		almost surely. To finish the proof, we note that by the previous analysis, if $\delta>0$, then we have
		\begin{equation*}
			S_i(T_k)\leq \frac{2}{\delta\Delta_i^2}\log(T_k)
		\end{equation*}
		almost surely for any big enough $k$. Now if $T_k< T\leq T_{k+1}$, then 
		\begin{align*}
			S_i(T)&=\frac{T_{k+1}-T}{T_{k+1}-T_k}S_i(T_k)+\frac{T-T_k}{T_{k+1}-T_k}S_i(T_{k+1})\\
			&\leq \frac{2}{\delta\Delta_i^2}\Big(\frac{T_{k+1}-T}{T_{k+1}-T_k}\log(T_k)+\frac{T-T_k}{T_{k+1}-T_k}\log(T_{k+1})\Big)\\
			&\leq \frac{2}{\delta\Delta_i^2}\log\Big(\frac{T_{k+1}-T}{T_{k+1}-T_k}T_k+\frac{T-T_k}{T_{k+1}-T_k}T_{k+1}\Big)\\
			&\leq \frac{2}{\delta\Delta_i^2}\log(T)
		\end{align*}
		where the second inequality follows from the concavity of $\log$. Finally, this analysis leads to
		\begin{equation*}
			\limsup_{T\rightarrow\infty}\frac{S_i(T)}{\log(T)}\leq \frac{2}{\Delta_i^2}
		\end{equation*}
		almost surely, since $\delta>0$ is an arbitrary rational number.
	\end{IEEEproof}
\end{theorem}
Using this theorem as the basis, we now show $B(T)=O(\log(T))$.

First of all, 
\begin{align*}
	\sum_{i=2}^IS_i(T_{k+1})-S_i(T_k)&\geq\sum_{i=2}^I\Prob(A_{T_k+1}=i|\Hp_{T_{k}})\Big(\frac{1}{\Prob_{2}(T_k)}-1\Big)\\
	&\geq 1-\sum_{i=2}^I\Prob(A_{T_k+1}=i|\Hp_{T_{k}})\\
	&=B(T_{k+1})-B(T_{k})-\sum_{i=2}^I\Prob(A_{T_k+1}=i|\Hp_{T_{k}})
\end{align*}
almost surely for any big enough $k$ by \eqref{inv}. Since $\sum_{i=2}^I\Prob(A_{T_k+1}=i|\Hp_{T_{k}})$ converges to 0 with probability 1, we have
\begin{equation*}
	\limsup_{j\rightarrow\infty}\frac{B(T_j)}{\sum_{i=2}^IS_i(T_{j})}\leq 1
\end{equation*}
almost surely by Stolz-Ces\`aro Theorem. In addition, by the definition of $B(T)$, we know that if $T_j\leq T<T_{j+1}$, then $B(T)=B(T_j)$. Combining this observation with the preceding analysis leads to
\begin{equation*}
	B(T)=B(T_j)
	\leq(1+\delta)\sum_{i=2}^IS_i(T_{j})
	\leq(1+\delta)\sum_{i=2}^IS_i(T)
\end{equation*}
almost surely for big enough $T$ and some $\delta>0$. Since $\delta$ is arbitrary, we have
\begin{equation*}
	\limsup_{T\rightarrow\infty}\frac{B(T)}{\sum_{i=2}^IS_i(T)}\leq 1
\end{equation*}
almost surely, which leads to
\begin{align*}
	\limsup_{T\rightarrow\infty}\frac{B(T)}{\log(T)}&=\limsup_{T\rightarrow\infty}\frac{\sum_{i=2}^IS_i(T)}{\log(T)}\frac{B(T)}{\sum_{i=2}^IS_i(T)}\\
	&\leq \sum_{i=2}^I\frac{2}{\Delta_i^2}
\end{align*}
with probability 1, where the last step follows from Theorem \ref{thm3}.

As for $R(T)$, by Theorem \ref{thm3} and \eqref{f2}, we have
\begin{align*}
	\limsup_{T\rightarrow\infty}\frac{R(T)}{\log(T)}&\leq\limsup_{T\rightarrow\infty}\sum_{i=2}^I\frac{R_i(T)}{\log(T)}\\
	&=\limsup_{T\rightarrow\infty}\sum_{i=2}^I\frac{R_i(T)}{S_i(T)}\frac{S_i(T)}{\log(T)}\\
	&\leq\sum_{i=2}^I\frac{2}{\Delta_i}
\end{align*}
almost surely. This analysis finishes the proof of Theorem \ref{thm6}.
\end{document}